\documentclass[pdflatex,sn-vancouver-num]{sn-jnl}

\usepackage[T1]{fontenc}
\usepackage[utf8]{inputenc}
\usepackage{lmodern}

\usepackage{graphicx}%
\usepackage{float}%
\usepackage{multirow}%
\usepackage{amsmath,amssymb,amsfonts}%
\usepackage{amsthm}%
\usepackage{mathrsfs}%
\usepackage[title]{appendix}%
\usepackage{xcolor}%
\usepackage{textcomp}%
\usepackage{manyfoot}%
\usepackage{booktabs}%
\usepackage{algorithm}%
\usepackage{amsmath}
\usepackage{algorithmicx}%
\usepackage{algpseudocode}%
\usepackage{listings}
\usepackage{booktabs}
\usepackage{array}
\usepackage{enumitem}
\setlist[itemize]{left=0pt..1em}

\usepackage{tcolorbox} 
\tcbuselibrary{listings, skins}

\usepackage[symbol]{footmisc}

\theoremstyle{thmstyleone}%
\theoremstyle{thmstyletwo}%

\theoremstyle{thmstylethree}%

\begin{document}

\title[VertAX: a differentiable vertex model for learning epithelial tissue mechanics]{\texttt{VertAX}: a differentiable vertex model for learning epithelial tissue mechanics}


\author[1]{\fnm{Alessandro} \sur{Pasqui}}

\author[1,2]{\fnm{Jim Martin} \sur{Catacora Ocana}}\equalcont{These authors contributed equally to this work.}
\author[1,2]{\fnm{Anshuman} \sur{Sinha}}\equalcont{These authors contributed equally to this work.}
\author[1,2]{\fnm{Matthieu} \sur{Perez}}\equalcont{These authors contributed equally to this work.}
\author[1,2]{\fnm{Fabrice} \sur{Delbary}}
\author[3,4]{\fnm{Giorgio} \sur{Gosti}}
\author[3,5]{\fnm{Mattia} \sur{Miotto}}
\author[3,5]{\fnm{Domenico} \sur{Caudo}}

\author*[6]{\fnm{Maxence} \sur{Ernoult}}\email{mernoult@google.com}
\author*[1,2]{\fnm{Hervé} \sur{Turlier}}\email{herve.turlier@cnrs.fr}

\affil[1]{Center for Interdisciplinary Research in Biology (CIRB), Collège de France, Université PSL, CNRS, INSERM, 75005 Paris, France}
\affil[2]{\textit{Current address:} Center for Integrative Biology (CBI), University of Toulouse, CNRS, Toulouse, France}
\affil[3]{Center for Life Nano \& Neuro Science, IIT, 00161, Rome, Italy}
\affil[4]{\textit{Current address:} Institute of Heritage Science, CNR, 00010, Montelibretti, Italy}
\affil[5]{Department of Physics, Università Sapienza, 00185, Rome, Italy}
\affil[6]{Google DeepMind\textsuperscript{\textdaggerdbl}}

\footnotetext[3]{Participated in an advisory capacity only.}

\abstract{
Epithelial tissues dynamically reshape through local mechanical interactions among cells, a process well captured by vertex models. Yet their many tunable parameters make inference and optimization challenging, motivating computational frameworks that flexibly model and learn tissue mechanics.
We introduce \texttt{VertAX}, a differentiable JAX-based framework for vertex-modeling of confluent epithelia. \texttt{VertAX} provides automatic differentiation, GPU acceleration, and end-to-end bilevel optimization for forward simulation, parameter inference, and inverse mechanical design. Users can define arbitrary energy and cost functions in pure Python, enabling seamless integration with machine-learning pipelines.
We demonstrate \texttt{VertAX} on three representative tasks: (i) forward modeling of tissue morphogenesis, (ii) mechanical parameter inference, and (iii) inverse design of tissue-scale behaviors. We benchmark three differentiation strategies—automatic differentiation, implicit differentiation, and equilibrium propagation—showing that the latter can approximate gradients using repeated forward, adjoint-free simulations alone, offering a simple route for extending inverse biophysical problems to non-differentiable simulators with limited additional engineering effort.
}

\keywords{vertex model, inverse problem, bilevel optimization, differentiable modeling, automatic differentiation, implicit differentiation, equilibrium propagatation, force inference, inverse design}


\maketitle

\section{Introduction}\label{introduction}

Epithelial tissues are dynamic physical systems that self-organize into complex structures, actively reshaping themselves during development, repair, and disease~\cite{guillot2013mechanics}. Understanding and predicting how mechanical forces drive this organization remains a central challenge at the intersection of developmental biology, physics, and engineering~\cite{goodwin2021mechanics}. Epithelia are a fundamental class of tissue in the body: they line internal organs and cavities and form protective barriers such as the skin. Structured as polarized monolayers of tightly packed cells, epithelia undergo collective shape changes and cell rearrangements orchestrated by local mechanical forces: cortical tension, cell–cell adhesion, and internal pressure~\cite{luciano2022appreciating}. Decoding these forces from observable tissue morphology, elucidating how they couple to cell fate decisions, and exploring their role in tissue-level organization are key questions in tissue morphogenesis. Addressing these questions requires efficient computational models that balance physical realism with compatibility to biological data, all while ensuring algorithmic efficiency.

A widely adopted framework for modeling epithelial mechanics in two dimensions is the vertex model, which represents cells as polygons and cell–cell junctions as edges that evolve under force balance~\cite{nagai1988vertex,bi2015density,Bi2016,yang2017correlating,giavazzi2018flocking,merkel2018geometrically,merkel2019minimal,sussman2020interplay}. This coarse-grained approach has proven effective in simulating diverse biological processes, including convergent extension~\cite{sknepnek2023generating,claussen2024geometric}, tissue fluidization~\cite{krajnc2018fluidization,yamamoto2022non} and sorting~\cite{sahu2020small}. Vertex models naturally lend themselves to energy-based formulations, where the mechanical equilibrium configuration of the tissue is determined by minimizing an effective energy function. Despite their success in theory and simulation, few publicly available implementations of vertex models exist—whether in C/C++~\cite{brakke1992surface,mirams2013chaste,sussman2017cellgpu,sego2023general}, or Python~\cite{theis2021tyssue,sarkar2024graph}—and are, for a larger part, not maintained. Many groups develop their own vertex-model implementations in diverse programming languages, often tailored to specific applications. When such implementations remain lab-specific rather than being released as reusable community tools, they can lead to fragmented efforts, duplicated work, and limited collective progress—unlike the collaborative open-source ecosystems that have flourished in computer science \citep{pedregosa2011scikit, wolf2019huggingface, blondel2022efficient}. \texttt{VertAX}, by contrast, is conceived as a public, generic implementation that others can reuse, extend, and build upon.
Finally, while differentiable programming frameworks such as PyTorch~\cite{paszke2017automatic} or Jax~\cite{bradbury2021jax} provide new opportunities for solving inverse problems, mechanical and parameter inference methods~\cite{ishihara2013comparative, ichbiah2023embryo, marin2023mapping} have so far relied on ``non-differentiable'' approaches (\emph{e.g.} C++ simulators), and areas such as inverse tissue design~\cite{deshpande2024engineering} are still emerging. 
Despite the availability of run-time \cite{hogan2014fast} and compile-time \cite{innes2019differentiable, vassilev2015clad} differentiation tools for C++, their high integration overhead persists as a bottleneck, necessitating more accessible differentiation strategies for high-performance simulators.

In parallel, machine learning has seen a rapid progress in methods for efficiently solving inverse and hyperparameter optimization problems, including for instance physics-informed neural networks (PINNs) \cite{raissi2019physics}, differentiable simulation \cite{de2018end,hu2019difftaichi,schoenholz2020jax} and model-based reinforcement learning \cite{lutter2021differentiable,tian2022real,moerland2023model,hatami2025robust}.
\emph{Bilevel optimization} (BO) \cite{sinha2017review,zucchet2022beyond} provides a general formalism to approach inverse problems involving nested optimization processes: a lower-level problem determining the system state (\emph{e.g.}, mechanical equilibrium) and an upper-level problem optimizing a parameter-dependent objective (\emph{e.g.}, matching experimental data or achieving a desired design). In essence, BO formulates a constrained optimization problem in which an outer loss is minimized with respect to model parameters under the constraint that an inner loss –defined by the physical model– is also minimized. Because vertex models are inherently energy-based, they can be naturally cast into this framework.

Here, we introduce \texttt{VertAX}, an end-to-end differentiable and open-source framework for vertex-based modeling of confluent epithelial tissues. Implemented in JAX~\cite{bradbury2021jax}, \texttt{VertAX} combines automatic differentiation, just-in-time compilation, and acceleration to enable fast and scalable simulations. A central contribution of this work is the unification of forward mechanics and inverse modeling into a bilevel optimization framework. Users can freely define custom energy (inner loss) and cost (outer loss) functions in pure Python, ensuring seamless integration with modern machine-learning workflows.

To enable gradient-based inverse modeling, \texttt{VertAX} implements and benchmarks three complementary differentiation strategies. \emph{Automatic differentiation} (AD)~\cite{baydin2018automatic} differentiates through the root finding algorithm solving for the inner optimization problem, \emph{i.e.} energy minimization.
\emph{Implicit differentiation} (ID)~\cite{dontchev2009implicit,lorraine2020optimizing,blondel2022efficient} avoids unrolling by instead differentiating through equilibrium directly. \emph{Equilibrium propagation} (EP)~\cite{scellier2017equilibrium,ernoult2019updates,scellier2019equivalence,zucchet2022beyond} 
emulates ID by estimating gradients from perturbed equilibria, requiring repeated executions of the same simulator.
We demonstrate the versatility of \texttt{VertAX} in solving inverse problems including parameter inference and inverse mechanical, morphological, or patterning design of epithelial tissues. Looking ahead, EP may introduce a novel approach to differentiability for legacy C++ codebases, enabling gradient computation by treating the simulator as a monolithic primal-only primitive rather than requiring intrusive internal instrumentation.

\section{Results}

\subsection{Forward vertex-based modeling in JAX}

At its simplest usage, \texttt{VertAX} serves as a ready-to-use, high-performance two-dimensional vertex model for forward mechanical simulations. Implemented as a Python module built on JAX~\cite{bradbury2021jax}, all core routines are just-in-time compiled to accelerate repeated optimization steps, and \texttt{VertAX} runs seamlessly on both CPUs and GPUs (see Methods for implementation details). The framework natively supports automatic differentiation (AD), allowing users to specify any phenomenological energy function as a standard Python function. \texttt{VertAX} then automatically minimizes this energy with respect to the vertex positions using either user-defined routines or any optimizer available in the \textit{Optax} library~\cite{deepmind2020jax}. \\

\begin{figure}[H]
  \centering
  \includegraphics[width=1.0\textwidth]{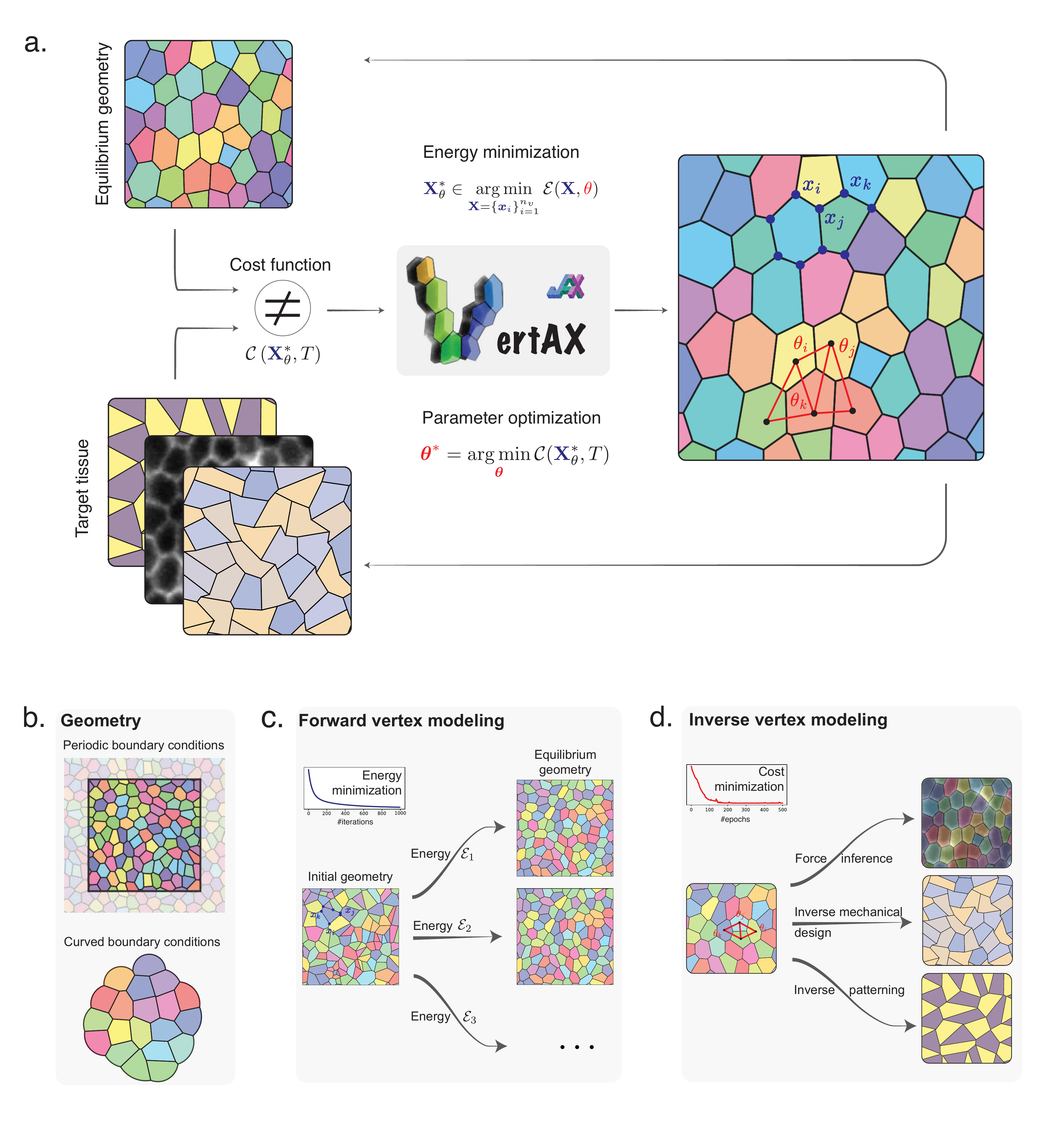}
\caption{\textbf{\texttt{VertAX} concept and forward/inverse modeling.}
\textbf{(a)}~In \texttt{VertAX}, tissue equilibrium is obtained by minimizing an energy~$\mathcal{E}(\mathbf{X},\boldsymbol{\theta})$ with respect to vertex positions~$\mathbf{X}$; in inverse problems, model parameters~$\boldsymbol{\theta}$ are optimized by minimizing a cost~$\mathcal{C}$ relative to a target.
\textbf{(b)}~Simulation modes with periodic or curved boundary conditions.
\textbf{(c)}~Forward modeling: different energy formulations produce different equilibrium geometries from the same initial state.
\textbf{(d)}~Inverse modeling applications: force inference, inverse mechanical design, and inverse patterning.}
    \label{fig:figure1}
\end{figure}

{\bfseries Energy functions.} Let $\mathcal{E}(\mathbf{X}, \boldsymbol{\theta})$ denote an energy function depending on the vertex positions $\mathbf{X} := \left\lbrace \mathbf{x}_i\right\rbrace_{i=1}^{n_v}$ and a set of model parameters $\boldsymbol{\theta} = \left\lbrace \theta_k\right\rbrace_{i=k}^{n_\theta}$ (Figure~\ref{fig:figure1}a). AD is used to compute the gradient of $\mathcal{E}$ with respect to $\mathbf{X}$, and this gradient is directly passed to the chosen optimizer to minimize the energy and compute the mechanical equilibrium configuration $\mathbf{X}^{*}_\theta \in \arg \min_{X} \mathcal{E}(\mathbf{X}, \boldsymbol{\theta})$, 
which obeys the stationarity condition 
$\nabla_{\mathbf{X}}\mathcal{E}(\mathbf{X}^{*}_\theta,\boldsymbol{\theta}) = 0$.

In this work, we consider two classical energy functions commonly used to describe epithelial mechanics in two dimensions~\cite{nagai2001dynamic,bi2015density,damavandi2025universality} (Figure~\ref{fig:figure1}b):
\begin{align}
\mathcal{E}_1(\mathbf{X},\theta) &= \sum_{\rm cell \ \alpha} \big(a_\alpha(\mathbf{X}) - 1\big)^2 + \sum_{\rm cell \ \alpha} \big(p_\alpha(\mathbf{X}) - p^0_\alpha\big)^2, \label{eq:Energy1}\\
\mathcal{E}_2(\mathbf{X},\theta) &= \frac{1}{2} K \sum_{\rm cell \ \alpha} \big(A_\alpha(\mathbf{X}) - A^0_\alpha\big)^2 + \sum_{\rm edge \ ij} \gamma_{ij} \ell_{ij}(\mathbf{X}). \label{eq:Energy2}
\end{align}
The first energy, $\mathcal{E}_1$, is a dimensionless formulation  defined on the set of cells $\{\alpha\}_{\alpha=1}^{n_c}$ and composed of elastic terms on rescaled cell areas $a_\alpha = A_\alpha/A_0$ (with $A_\alpha$ the area of cell $\alpha$ and $A_0$ the target area) and shape factors $p_\alpha = P_\alpha/\sqrt{A_0}$ (with $P_\alpha$ the perimeter of cell $\alpha$), using a uniform target rescaled area $a^0=1$ and heterogeneous target shape factors $p^0_\alpha$ that may vary across cells, and considered as model parameters $\theta$. The second energy, $\mathcal{E}_2$, includes an elastic penalty on cell areas (with elastic constant $K$) and a line tension term acting on each edge of length $\ell_{ij}$ (between vertices $i$ and $j$) weighted by its corresponding line tension coefficient $\gamma_{ij}$, considered as model parameters $\theta$. Other phenomenological energy-based vertex models~\cite{lange2025vertex} can be implemented just as easily, provided they are explicit functions of vertex positions $\mathbf{X}$ and tissue topology. Model parameters can be defined per cell, edge, or vertex, allowing the representation of heterogeneous mechanical properties within a single tissue. \\

{\bfseries Boundary modes.} \texttt{VertAX} supports two main simulation boundary modes: a periodic mode, suitable for modeling bulk tissue dynamics without explicit boundaries, and a bounded mode, which explicitly represents curved tissue interfaces and interface forces (Figure~\ref{fig:figure1}b). In the periodic mode, the tissue is enclosed within a two-dimensional simulation box with periodic boundary conditions, such that when a vertex crosses one boundary, it is automatically reintroduced on the opposite side to preserve topological continuity. Periodic vertex configurations can be initialized in two ways: either as random periodic Voronoi tessellations generated from user-defined seed points, or directly from segmented tissue images—such as two-dimensional semantic masks produced by \textit{cellpose}~\cite{stringer2021cellpose}. In the bounded mode, tissue edges are explicitly defined as circular arcs whose curvature is treated as an additional degree of freedom (via an angle defined per edge), allowing the model to satisfy mechanical equilibrium at outer tricellular junctions. When the distance between two vertices falls below a user-defined threshold, a neighbor exchange (T1 topological transition) is triggered if it results in a reduction of the total energy. We refer the reader to the Methods section for details on the data structure, implementation of periodic and bounded boundaries, and T1 transitions.

\subsection{Bilevel optimization for inverse modeling}

Beyond classical forward simulations, the core novelty of \texttt{VertAX} lies in its ability to seamlessly solve inverse problems within the same computational framework (Figure~\ref{fig:figure1}d). In these problems, the objective is not only to simulate tissue mechanics but to optimize model parameters—such as shape factors, line tensions, or elastic moduli—so that the predicted equilibrium configuration reproduces or achieves a desired outcome. This setting can be naturally formulated as a bilevel optimization problem, in which the inner (lower-level) problem determines the tissue state by minimizing a mechanical energy function $\mathcal{E}$, and the outer (upper-level) problem adjusts model parameters to minimize a user-defined cost function $\mathcal{C}$ that quantifies the mismatch to data or deviation from a design target:
\begin{equation}
\min_{\boldsymbol{\theta}} \mathcal{C}(\mathbf{X}^*_{\boldsymbol{\theta}},\theta)
\quad \text{s.t.} \quad
\mathbf{X}^{*}_{\boldsymbol{\theta}} \in \arg\min_{\mathbf{X}} \mathcal{E}(\mathbf{X}, \boldsymbol{\theta}).
\label{def:bo-problem}
\end{equation}

A central computational challenge in bilevel optimization is differentiating through the implicit or iterative solution of the inner problem $\mathbf{X}^{*}_{\boldsymbol{\theta}}$. To address this, \texttt{VertAX} implements three complementary hypergradient strategies:

\textbf{\emph{Automatic differentiation (AD)}:}~~
AD differentiates through the full inner minimization trajectory by applying the chain rule across all relaxation steps \cite{baydin2018automatic}. This yields exact hypergradients, but requires storing the unrolled computational graph in memory.

\textbf{\emph{Implicit differentiation (ID):}}~~
ID computes exact hypergradients directly at equilibrium using the implicit function theorem, avoiding trajectory unrolling and substantially reducing memory usage \cite{dontchev2009implicit, lorraine2020optimizing, blondel2022efficient}. In \texttt{VertAX}, we consider both a sensitivity mode (Sen-ID) and an adjoint-state mode (Adj-ID) \cite{plessix2006review}. In practice, however, ID requires solving a linear system involving the mechanical Hessian, which can become ill-conditioned near marginally stable states \cite{zucchet2022beyond}.

\textbf{\emph{Equilibrium propagation (EP)}:}~~
EP estimates hypergradients from small perturbations around equilibrium, without explicit backpropagation through the solver \cite{scellier2017equilibrium, ernoult2019updates, zucchet2022beyond}. In \texttt{VertAX}, we use a centered second-order estimator based on symmetric perturbations \cite{laborieux2021scaling, scellier2023energy}. EP is highly memory-efficient and remains applicable even when the simulation engine is only approximately differentiable\footnote{Note that EP is distinct from Zeroth-Order (ZO) perturbation methods which also employ several repeated executions of the same program using random perturbations of the parameters \cite{spall1998overview}, which are plagued with high-variance gradient estimates. In contrast, EP yields unbiased and variance-free gradient estimates.}.
\\

Despite their complementary advantages, these three strategies have not previously been benchmarked side-by-side within a unified vertex-modeling framework.

\subsubsection{Benchmarking bilevel optimization approaches on vertex models}

To quantitatively assess the accuracy and efficiency of the three differentiation strategies implemented in \texttt{VertAX}, we benchmarked reverse AD, Adj-ID, Sen-ID, and EP on a synthetic inverse problem constructed from 
the energy formulation $\mathcal{E}_1$. Ground-truth configurations were first generated by forward minimization of the respective energy functions, using respectively cell-specific target shape factors $p^0_\alpha$ 
sampled from Gaussian distributions.
The inverse problem then aimed to recover these parameters by minimizing, as cost function, the mean-squared error between predicted and ground-truth vertex positions (further called vertex-to-vertex, or v2v, cost function): 
\begin{equation}
\mathcal{C}(\mathbf{X}, \mathbf{X}_{\text{GT}}) = \frac{1}{2N}\sum_{k=1}^N \|\mathbf{x}_k - \mathbf{x}_{\text{GT},k}\|^2.
\end{equation}

Initial conditions were generated by re-sampling the tunable parameters from the same prior distributions, yielding distinct equilibrium geometries while preserving the number of degrees of freedom. Inner mechanical relaxation was performed by gradient descent, and outer parameter updates were carried out with ADAM and Nesterov momentum. Unless otherwise stated, all methods were run with identical hyperparameters. Performance was assessed using a mean relative L1 error and a Pearson correlation metric between inferred and ground-truth parameters, and the final value of the cost function.

\begin{figure}[H]
  \centering
  \includegraphics[width=1.0\textwidth]{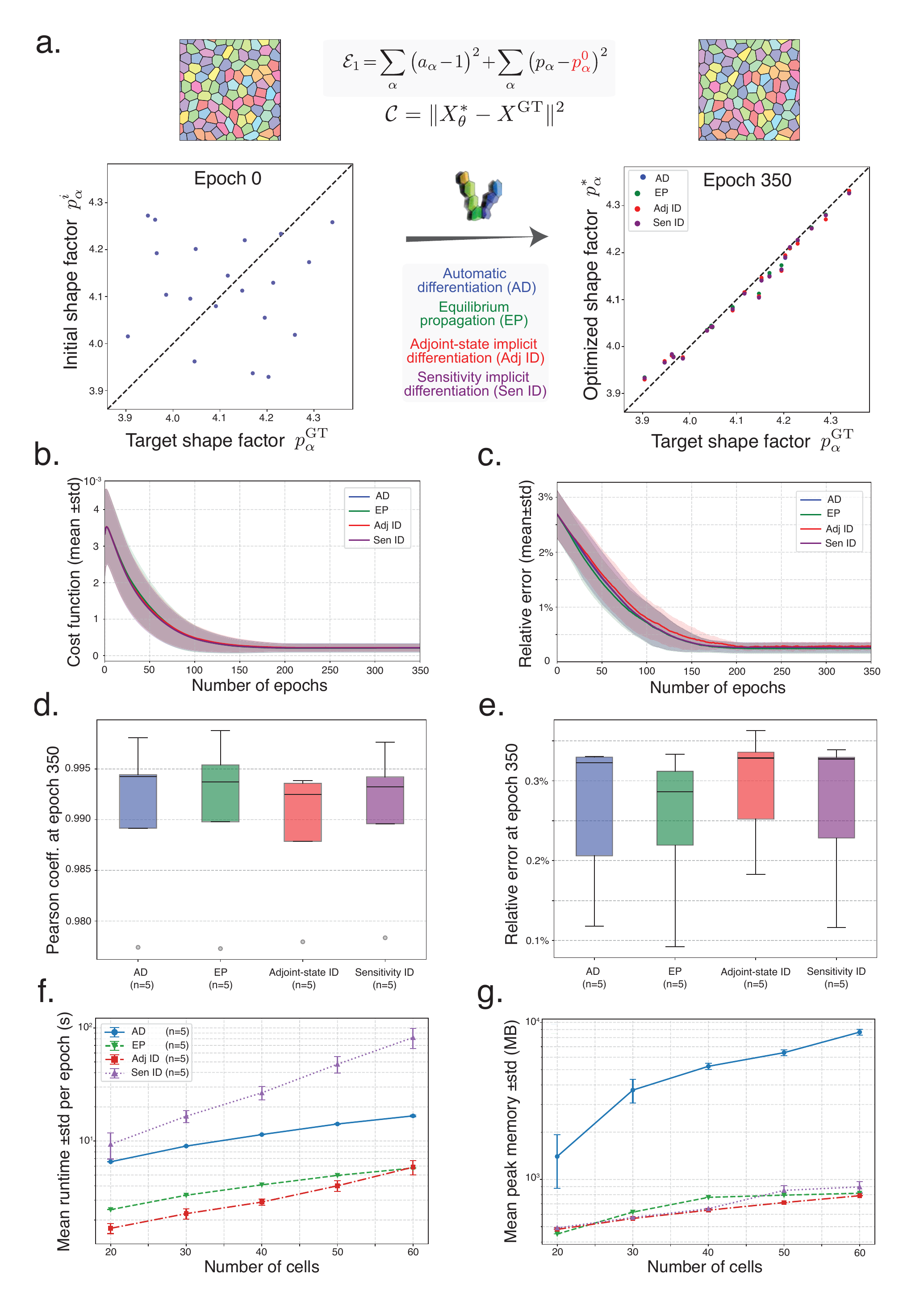}
  \caption{\textbf{Benchmarking bilevel optimization strategies for mechanical parameter inference in \texttt{VertAX}.}
\textbf{(a)}~Synthetic benchmark for inference of target shape factors~$p^0_\alpha$. Left, inferred versus ground-truth values at epoch~0; right, inferred versus ground-truth values at epoch~350. Results are shown for AD (blue), EP (green), Adj-ID (red), and Sen-ID (purple). The dashed diagonal indicates perfect agreement. The benchmark uses the energy
$\mathcal{E}_1$ and the v2v cost.
\textbf{(b, c)}~Evolution over training epochs of the mean cost function~$\mathcal{C}$ (b, lower is better) and mean relative error (c, lower is better).
\textbf{(d, e)}~Box plots of the Pearson correlation coefficient~(d, higher is better) and relative error~(e, lower is better) at epoch~350.
\textbf{(f, g)}~Mean runtime per epoch~(f, lower is better; log scale) and peak memory usage~(g, lower is better; log scale) as a function of cell number.
\textbf{Experimental details:} Data in (b--g) show mean $\pm$ s.d. over $n=5$ independent replicates. Unless otherwise specified in (f, g), tissues contained $N=20$ cells with $L_{\mathrm{box}}=\sqrt{N}$. Outer-loop optimization used ADAM with learning rate $10^{-3}$, and inner-loop energy minimization used gradient descent with learning rate $10^{-2}$. Inner-loop stopping parameters were a maximum of $10^3$ iterations, tolerance $10^{-3}$, patience $10$, and minimum T1 distance $5\times10^{-3}$. For EP, the optimal finite difference parameter $\beta$ was pre-selected to maximize performance (Extended Figure 1).
}
\label{fig:figure2}
\end{figure}

Across experiments averaged over five replicates, all four methods successfully recovered the target parameters (Figure~\ref{fig:figure2}a). Convergence of the cost and relative error occurred within approximately 200 epochs (Figure~\ref{fig:figure2}b,c). By epoch 350, the methods achieved comparable Pearson correlations and relative errors (Figure~\ref{fig:figure2}d,e), with EP showing a slight performance advantage.

We next compared the methods in terms of mean runtime per epoch and mean peak memory usage as the number of cells increased. For runtime, Adj-ID was marginally faster than EP, followed by AD and Sen-ID, with all methods remaining within one order of magnitude of each other (Figure~\ref{fig:figure2}f). In contrast, AD consistently required about tenfold more memory than EP and the two ID-based methods (Figure~\ref{fig:figure2}g). Taken together, these results indicate that Adj-ID and EP offer the best overall trade-off between reconstruction accuracy, computational speed, and memory efficiency.


\subsection{Vertex model parameter inference}

\subsubsection{Topological cost functions unlock robust parameter inference}

In our earlier benchmarks evaluating gradient estimation strategies (Figure~\ref{fig:figure2}), we intentionally initialized the predicted tissue relatively close to the target configuration. This precaution ensured that the physical tissue would naturally relax into the correct topological state, allowing us to isolate and evaluate the computational performance of AD, ID, and EP using a simple geometric vertex-to-vertex (v2v) outer loss. 

However, in realistic parameter inference scenarios, the initial guess and target configurations may often reside in different topological states. The mechanical energy landscape of vertex models is notoriously rugged due to the discrete nature of cellular connectivity \cite{bi2015density, barton2017active}. Because the inner optimization process (simulating physical relaxation) deterministically halts at the nearest local energy minimum, the tissue cannot spontaneously rewire. To reach the correct target state, the outer optimization loop must temporarily adjust the mechanical parameters to actively drive the system through a series of energy barriers and cell neighbor exchanges (T1 transitions).

This demonstrates that choosing the right outer cost function is a non-trivial and critical task. We found that a purely geometric v2v distance (Figure~\ref{fig:figure3}a, left) generally fails to infer the correct biophysical parameters when T1 topological barriers are present (Figure~\ref{fig:figure3}b-c). Because the v2v loss solely penalizes spatial deviations of the vertices, it lacks the structural information necessary to generate the parameter gradients required to trigger T1 transitions. Consequently, the outer optimization becomes trapped in a sub-optimal local minimum in the parameter space, leaving a high residual topological error.

To overcome this fundamental limitation, we incorporated the Index-Aware Structural (IAS) loss (Figure~\ref{fig:figure3}a, right), a differentiable topological metric rooted in optimal transport \cite{vayer2019optimal, cuturi2013sinkhorn}. Mathematically, the IAS loss computes a continuous structural signature for each cell based on the dual adjacency matrix $A$ of the tissue. We define a continuous structure matrix $S$ as:
\begin{equation}
    S = 2I - A - \frac{1}{2}A^2
\end{equation}
where the $A^2$ term captures the 2-hop topological neighborhood of each cell, providing a wider spatial ``receptive field'' smoothing discrete, step-like barriers associated with topological rewiring. This allows the optimizer to anticipate and guide T1 transitions before they become geometrically favorable. 

Remarkably, incorporating just a 1\% weighting of the IAS loss ($0.99~\text{v2v} + 0.01~\text{IAS}$) allowed the bilevel optimizer to confidently break local symmetries, successfully guiding the tissue through the required T1 transitions. This fusion resulted in largely improved parameter retrieval (mean relative error approaching $0\%$ and Pearson coefficient approaching $1.0$) and effectively reduced the relative topological error (Figure~\ref{fig:figure3}d-e).

\subsubsection{Inference of mechanical parameters from microscopy image}

\begin{figure}[H]
\centering
\includegraphics[width=0.85\textwidth]{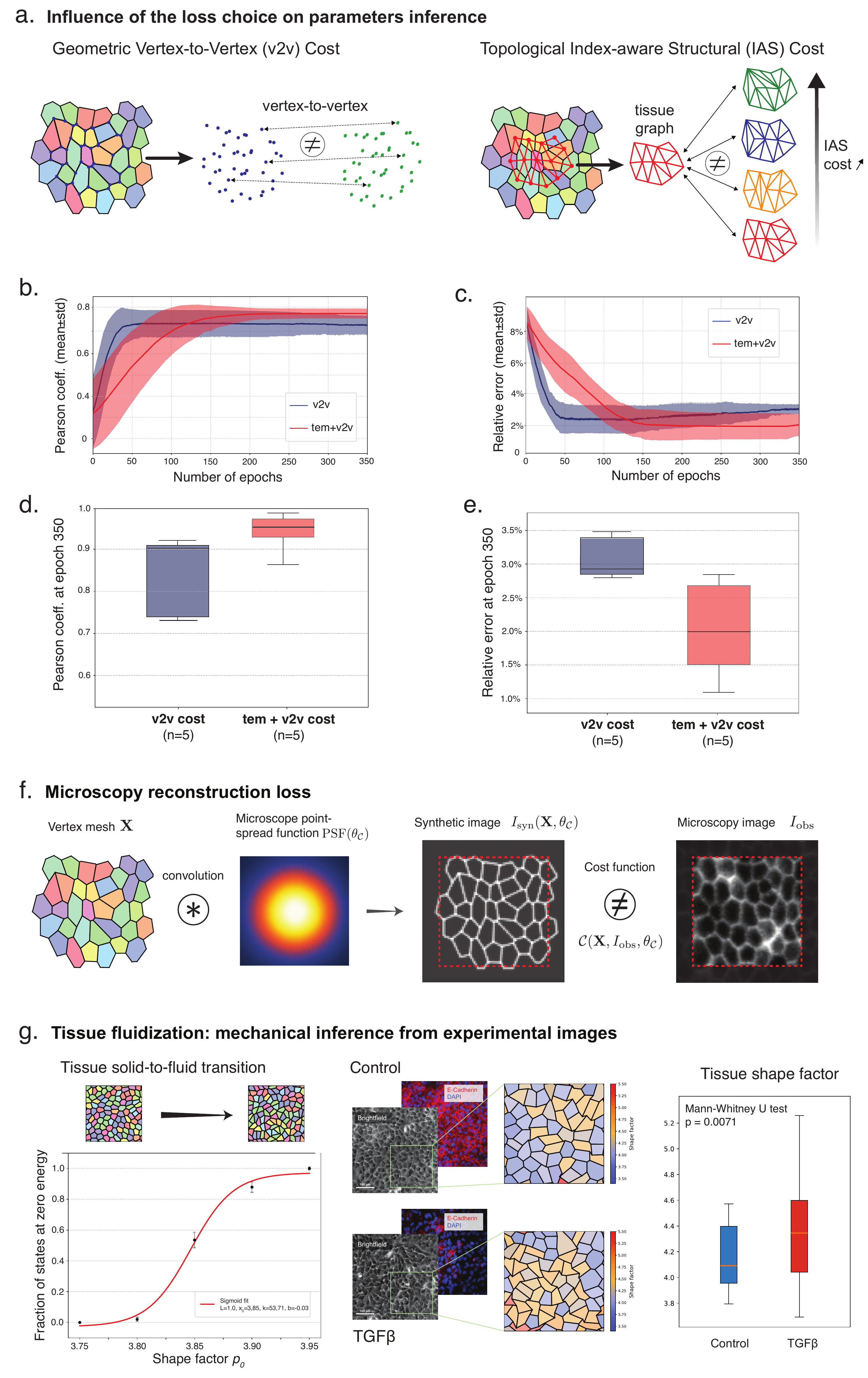}
\caption{
\textbf{Structural and image-based losses for inverse vertex modeling in \texttt{VertAX}.}
\textbf{(a)}~Comparison of two outer-loss functions for bilevel parameter inference from tissue configurations initialized far from the target: a geometric vertex-to-vertex (v2v) cost and an index-aware structural (IAS) cost based on the tissue graph topology.
\textbf{(b, c)}~Evolution over training epochs of the Pearson correlation coefficient between inferred and ground-truth parameters (b) and of the relative error (c) for inference with the v2v cost alone or with a combined loss (IAS+v2v).
\textbf{(d, e)}~Box plots of the Pearson correlation coefficient (d) and relative error (e) at epoch~350 across $n=5$ independent initializations.
\textbf{(f)}~Schematic of the microscopy reconstruction loss. The polygonal vertex mesh~$\mathbf{X}$ is convolved with a microscope point-spread function~$\mathrm{PSF}(\theta_{\mathcal C})$ to generate a synthetic image~$I_{\mathrm{syn}}(\mathbf{X},\theta_{\mathcal C})$. The outer cost $\mathcal{C}(\mathbf{X},I_{\mathrm{obs}},\theta_{\mathcal C})$ is defined from the discrepancy between this synthetic image and the microscopy image~$I_{\mathrm{obs}}$.
\textbf{(g)}~Application to tissue fluidization inference from experimental images. Left: fraction of zero-energy states as a function of the target shape factor~$p^0$ from forward simulations, with sigmoid fit shown in red ($n=100$ independent simulations of 100-cell tissues). Center: representative control and TGF$\beta$-treated A549 monolayers, with brightfield and E-cadherin/DAPI images, together with the inferred equilibrium mesh initialized from a Voronoi tessellation of Cellpose-segmented nuclei and optimized using the microscopy reconstruction loss. Right: box plots of inferred tissue shape factors for control and TGF$\beta$ conditions (Mann--Whitney $U$ test, $p = 0.0071$).
}
\label{fig:figure3}
\end{figure}

Beyond benchmark geometries, we demonstrated the biological relevance of our inference pipeline by applying it to tissue fluidization (Figure ~\ref{fig:figure3}g).

As an initial validation, forward simulations confirmed that \texttt{VertAX} correctly captures the classical rigidity transition from a solid-like to a fluid-like phase as the target shape factor $p^0$ increases \citep{sussman2020interplay} (Figure ~\ref{fig:figure3}g, left).

We next extended the bilevel inference framework to real microscopy data by implementing an image-mesh outer cost, which demonstrated its efficiency for active mesh-based segmentation in 3 dimensions~\cite{ichbiah2025inverse}. In this setting, the cost function quantifies the discrepancy between a simulated fluorescence image and the corresponding experimental microscopy image. The simulated image $I_A(\mathbf{X})$ is obtained by convolving the cell-boundary polygonal mesh $\mathbf{X}$ with a Gaussian point spread function (PSF) (Figure~\ref{fig:figure3}f), mimicking the optical blurring of fluorescence microscopy. The image-based cost is then defined as a pixelwise mean squared deviation between the artificial and experimental images:
\begin{equation}
\mathcal{C}(I_A, I_{\mathrm{exp}}) = \frac{1}{2M}\sum_{k=1}^{M}\big(I_A(\mathbf{r}_k) - I_{\mathrm{exp}}(\mathbf{r}_k)\big)^2,
\end{equation}
where $M$ is the number of pixels and $\mathbf{r}_k$ their coordinates. Implemented in JAX with Fast Fourier Transform, this differentiable cost enables direct gradient-based adjustment of the mesh to align the simulated mesh geometry with observed cell boundaries.
Used as an outer cost function acting on the equilibrium vertex configuration $\mathbf{X}^*_\theta$, the bilevel-optimization problem transforms into a vertex model parameter inference directly from microscopy images. 

To demonstrate its biological applicability, we deployed this pipeline on A549 epithelial monolayers under standard (CTRL) and TGF-$\beta$-treated growing conditions. TGF-$\beta$ induces an epithelial-to-mesenchymal transition (EMT), which shifts epithelial tissues toward a more fluid-like, unjammed state and is associated with reduced adherens junction integrity, including decreased E-cadherin expression \cite{oswald2017jamming} (Figure~\ref{fig:figure3}g, center). We used bilevel optimization to infer cell-level target shape factors $p^0_\alpha$ by minimizing the image--mesh cost against the experimental images, excluding boundary cells to limit edge effects. Initial meshes were generated from Voronoi tessellations seeded by Cellpose-extracted nuclei \cite{stringer2021cellpose}. The inferred shape factors were consistently higher in TGF-$\beta$-treated tissues than in CTRL tissues, in line with increased tissue fluidization (Figure~\ref{fig:figure3}g, right). This approach demonstrates how inferring mechanical parameters directly from microscopy images may inform about the mechanical state of a tissue while bypassing the noise-prone bottleneck of explicit cell segmentation.

\subsection{Inverse design of epithelial tissues}

\begin{figure}[H]
\centering
\includegraphics[width=1.0\textwidth]{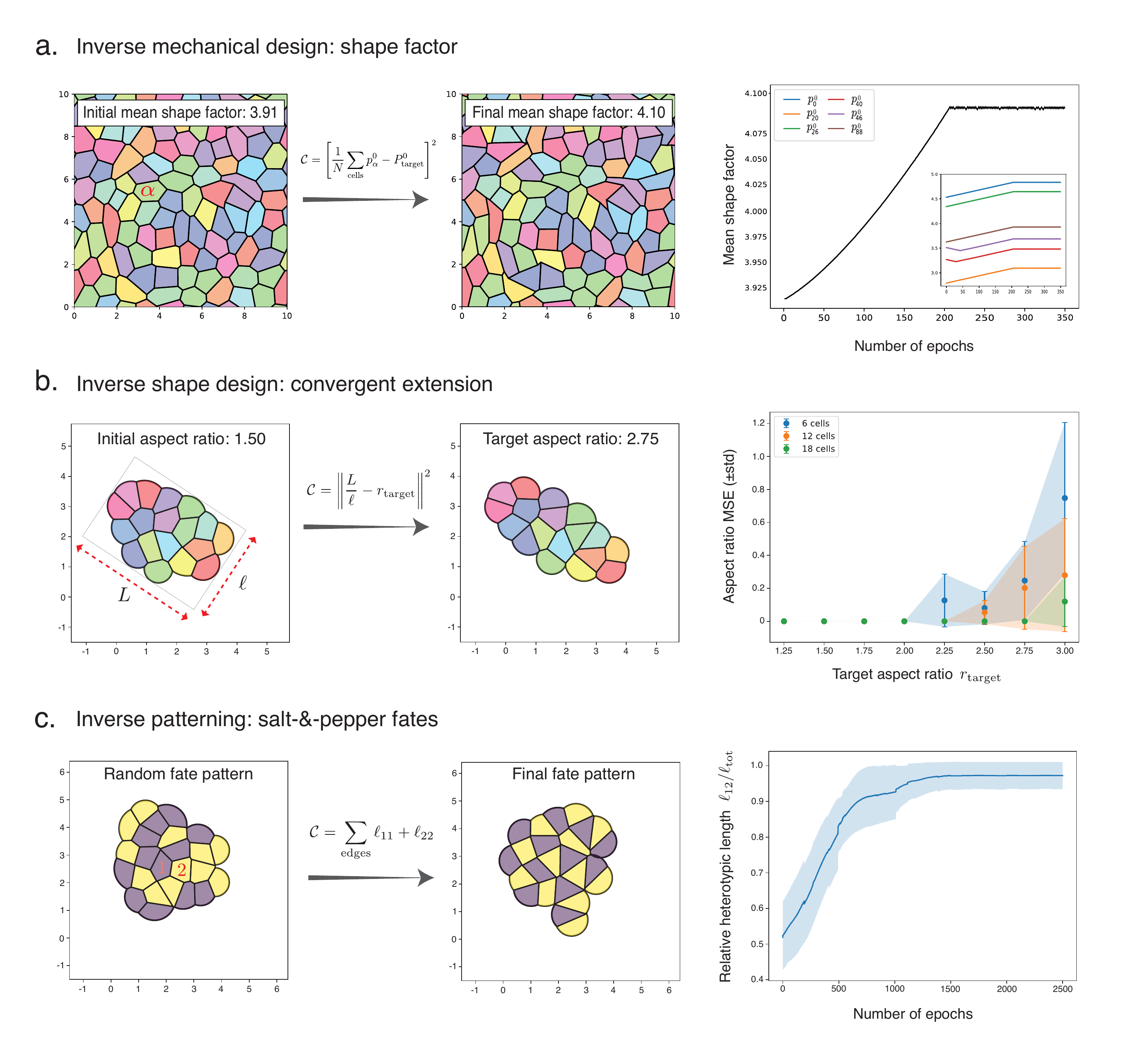}
\caption{\textbf{Inverse design of epithelial tissues with \texttt{VertAX}.}
\textbf{(a)}~Inverse mechanical design of target shape factor using a cost $\mathcal{C}$ that penalizes deviations of the mean equilibrium shape factor from $P^0_{\mathrm{target}}=4.10$. Left and center: initial and final equilibrium configurations. Right: epoch-wise evolution of the mean shape factor and of the individual target shape factors $p^0_\alpha$ (inset, early-epoch dynamics).
\textbf{(b)}~Inverse morphological design of tissue elongation using a cost $\mathcal{C}$ based on the squared deviation of the aspect ratio $r=L/\ell$ from a target value $r_{\mathrm{target}}$. Left and center: initial ($r = 1.50$) and final ($r = 2.75$) tissue configurations. Right: aspect-ratio mean squared error ($\pm$std) versus $r_{\mathrm{target}}$ for tissues of 6, 12, and 18 cells.
\textbf{(c)}~Inverse patterning design using a cost $\mathcal{C}=\sum_{\mathrm{edges}}\ell_{11}+\ell_{22}$ to favor heterotypic interfaces. Left and center: initial random and final optimized fate patterns. Right: mean $\pm$ std of the relative heterotypic interface length $\ell_{12}/\ell_{\mathrm{tot}}$ over epochs ($n=5$ replicates).
}
\label{fig:figure4}
\end{figure}

Inverse-design problems are of growing interest because they link mechanistic models of morphogenesis to the rational design of multicellular systems with prescribed shapes, structures, or collective behaviors \cite{deshpande2024engineering, velazquez2018programming}. Beyond parameter inference from experimental data, the bilevel optimization framework implemented in \texttt{VertAX} naturally extends to such problems, where the objective is not to recover parameters from observations, but to identify parameter sets that drive the tissue toward a prescribed emergent configuration or behavior. In this \emph{inverse design} setting, the cost function is no longer defined as a discrepancy between predicted and observed geometric and topological configurations, but instead evaluates a tissue-scale observable $\mathcal{O}(\mathbf{X}^*_{\boldsymbol{\theta}})$ measuring how closely the equilibrium configuration achieves a desired macroscopic property:
\begin{equation}
    \min_{\boldsymbol{\theta}}\; \mathcal{C}\!\left(\mathcal{O}(\mathbf{X}^*_{\boldsymbol{\theta}})\right)
    \quad \text{s.t.} \quad
    \mathbf{X}^*_{\boldsymbol{\theta}} \in \arg\min_{\mathbf{X}}\,\mathcal{E}(\mathbf{X},\boldsymbol{\theta}).
\end{equation}
Because users may define $\mathcal{C}$ as an arbitrary Python function of the equilibrium state, \texttt{VertAX} can target mechanical, morphological, or patterning objectives within a unified computational workflow. We demonstrate this capability on three illustrative inverse design tasks, each formulated as a bilevel optimization problem and summarized in Figure~\ref{fig:figure4}.

\paragraph{\emph{Inverse mechanical design.}}
In the first task, the energy function $\mathcal{E}_1$ governs tissue mechanics, and we sought to identify per-cell target shape factors $p^0_\alpha$ such that the mean equilibrium shape factor of the tissue matches a prescribed target value $P^0_\mathrm{target}$. The cost function was defined as
\begin{equation}
    \mathcal{C} = \left(\frac{1}{N}\sum_\alpha p^0_\alpha - P^0_\mathrm{target}\right)^2,
\end{equation}
penalizing deviations of the mean target shape factor from the desired value. Starting from an initial tissue with a mean shape factor of $3.91$, optimization successfully drove the tissue toward the target value of $4.10$ within approximately $200$ epochs. Individual per-cell shape factors $p^0_\alpha$ converge uniformly to distinct values that collectively realize the prescribed mean (Figure~\ref{fig:figure4}a, right), illustrating the ability of \texttt{VertAX} to distribute mechanical design targets across heterogeneous cell populations.

\paragraph{\emph{Inverse morphological design: convergent extension.}}
In the second task, we targeted the global shape of a bounded tissue cluster whose mechanics are governed by $\mathcal{E}_2$, specifically its aspect ratio $r = L/\ell$, defined as the ratio of the tissue's long to short axis. The cost function
\begin{equation}
    \mathcal{C} = \left\|\frac{L}{\ell} - r_\mathrm{target}\right\|^2
\end{equation}
penalizes deviations of the equilibrium aspect ratio from a prescribed target $r_\mathrm{target}$, mimicking the tissue-elongation process known as convergent extension~\citep{wallingford2002convergent}. Starting from an initial aspect ratio of $1.50$, \texttt{VertAX} successfully elongated the tissue toward targets up to $r_\mathrm{target} \approx 2.25$ with low mean-squared error across tissue sizes of $6$, $12$, and $18$ cells (Figure~\ref{fig:figure4}b, right). For larger targets ($r_\mathrm{target} \gtrsim 2.50$), the mean-squared error increased and exhibited greater variability, reflecting the growing difficulty of realizing highly anisotropic configurations within the mechanical constraints of the vertex model. Smaller tissues were found to be more sensitive to this effect, likely due to their reduced number of degrees of freedom.

\paragraph{\emph{Inverse patterning design.}}
In the third task, we addressed a patterning problem in which two cell types (labeled $1$ and $2$) are assigned heterogeneous line tension coefficients $\gamma_{ij}$, and the objective is to promote a salt-and-pepper fate arrangement---a checkerboard-like configuration in which cells of the same type avoid contact. The cost function was defined as the total length of homotypic interfaces:
\begin{equation}
\mathcal{C} = \sum_\mathrm{edges} \ell_{11} + \ell_{22},
\end{equation}
where $\ell_{11}$ and $\ell_{22}$ denote the total lengths of homotypic interfaces shared between cells of type $1$ and type $1$, and of type $2$ and type $2$, respectively. We denote by $\ell_{12}$ the total length of heterotypic interfaces between cells of type $1$ and type $2$, and by $\ell_{\mathrm{tot}}=\ell_{11}+\ell_{22}+\ell_{12}$ the total interface length in the tissue. Minimizing $\mathcal{C}$ thus drives the system toward configurations in which heterotypic contacts dominate. Starting from a random fate assignment, optimization rapidly increased the relative heterotypic interface length $\ell_{12}/\ell_\mathrm{tot}$ from approximately $0.5$ to near $1.0$ within $500$ epochs, reaching a stable salt-and-pepper pattern by epoch ${\sim}1000$ (Figure~\ref{fig:figure4}c, right).\\

Taken together, these three tasks illustrate the versatility of the \texttt{VertAX} inverse design framework: by simply specifying a cost function encoding the desired tissue-level behavior, diverse mechanical, morphological, and patterning design objectives can be addressed within a unified bilevel optimization workflow.

\section{Discussion}

In this work, we introduced \texttt{VertAX}, a fully differentiable framework that natively casts epithelial tissue mechanics as a bilevel optimization (BO) problem. By bridging classical vertex models with the modern JAX ecosystem, \texttt{VertAX} transforms the simulation engine from a pure forward-modeling tool into a versatile platform for inverse parameter inference and morphogenetic design.

Our benchmarking of gradient estimation strategies (Figure~\ref{fig:figure2}) provides critical insights for computational physics. While Automatic Differentiation (AD) yielded the fastest convergence, Equilibrium Propagation (EP) proved highly competitive, requiring only modest tuning of its hyperparameters. This demonstrates that EP is a powerful alternative capable of extending bilevel optimization to legacy, non natively differentiable simulators where implementing exact AD would require substantial engineering. Importantly, because EP estimates gradients from perturbed equilibria, it only requires the energy function to be lower-bounded. This is particularly attractive in vertex models, where exact physical equilibrium conditions are hard to reach, often yielding indefinite Hessians. Conversely, while Implicit Differentiation (ID) is theoretically exact and as memory-efficient as EP, but we found it sensitive to incomplete inner-loop energy minimization: ID necessitates robust least-squares linear solvers to handle indefinite Hessians and may be slower to converge \citep{domke2012generic}.

Beyond synthetic benchmarks, we demonstrated that \texttt{VertAX} can bridge abstract physical models with experimental realities. 
By integrating a differentiable optical cost function, we successfully inferred mechanical parameters—and distinguished physical states like TGF-$\beta$-induced tissue fluidization—directly from real microscopy images, circumventing the traditional bottleneck of heuristic cell segmentation (Figure~\ref{fig:figure3}).
Furthermore, we showcased the potential of this framework for synthetic morphology \textit{in silico}: by optimizing local cell-scale parameters against macroscopic target costs, we successfully programmed complex tissue-level behaviors, including convergent extension and fate patterning (Figure~\ref{fig:figure4}).

Crucially, our computational architecture provides a rigorous mathematical sandbox for exploring the emerging paradigms of ``physical learning'' \cite{stern2021supervised} and ``double optimization'' \cite{arzash2025rigidity} in living matter. In these theoretical proposals, physical networks dynamically adapt their local mechanical rules to minimize a global stress or target function. \texttt{VertAX} could be seemlessly extended to benchmark these algorithms alongside Equilibrium Propagation \cite{scellier2023energy}. From this perspective, EP is not seen here as merely a physical learning rule used for the exploration of novel computational paradigms, but as a tool which can be readily leveraged in software for bilevel optimization for vertex models. 


Ultimately, \texttt{VertAX} fills a critical gap between computational biophysics and differentiable programming. By enabling the non-invasive inference of mechanical parameters \textit{in vivo} and the programmable inverse design of active tissues, we anticipate this open-source framework will broadly empower both experimentalists and theorists studying the physical principles of morphogenesis.

\bmhead{Supplementary information}
A detailed description of the VertAX framework is provided in the accompanying Supplementary Information file.

\bmhead{Acknowledgements}
The authors wish to thank all members of the Turlier team for insightful discussions. GG, MM and DC thank the Imaging Facility at Center for Life Nano \& Neuro Science, Istituto Italiano di Tecnologia.

\bmhead{Funding}
This project received funding from the European Union’s Horizon 2020 research and innovation programme under the European Research Council grant agreement no. 949267 (H.T.) and under the Marie Skłodowska-Curie grant agreement No. 945304 – Cofund AI4theSciences, hosted by PSL University (A.P. and H.T.). M.M. acknowledges funding from the Italian Ministero dell’Università e della Ricerca, Project FIS2023-02957 (CUP B53C24009530001).

\bmhead{Competing interests}
The authors declare no competing interests.

\bmhead{Consent for publication}
All authors have given their consent for publication.

\bmhead{Data availability}
The datasets generated and analyzed in this study will be deposited in a public repository prior to publication. Accession details will be provided upon acceptance.

\bmhead{Code availability}
To ensure accessibility, VertAX is made freely available as open-source framework. The full codebase is hosted on GitHub: \href{https://github.com/VirtualEmbryo/VertAX}{https://github.com/VirtualEmbryo/VertAX}.

\bmhead{Author contribution}
H.T. conceived and supervised the project, and secured funding. H.T., M.E. and A.P. conceived the methodology. M.E. supervised the bilevel optimization implementation and analysis. A.P. developed the periodic VertAX codebase, with contributions from F.D. for the mesh-image cost function and implicit differentiation, J.M.C.O. for the bounded mode and A.S. for the topological cost. H.T., M.E., G.G. and M.M. provided guidance during the implementation steps of the codebase. A.P. performed the bilevel optimization analyses. A.S. performed the outer cost influence analysis. J.M.C.O. performed the inverse design analysis. D.C. performed the cell biology experiment. D.C., G.G., and M.M. performed microscopy experiments. A.P. and D.C. performed parameter inference analysis on the acquired data. H.T. conceptualized the manuscript and figures, which were created by A.P., J.M.C.O and A.S. H.T. wrote the manuscript with A.P. and M.E. All authors reviewed and approved the manuscript.

\section*{Methods}

\subsection*{Vertex model formulation}

The tissue is represented as a planar graph comprising $N$ vertices (nodes), where each face (polygon) represents a cell $\alpha$. The mechanical state of the tissue is governed by an energy functional that dictates the equilibrium configuration of the vertices. \texttt{VertAX} supports multiple energy formulations. For instance, a standard formulation $\mathcal{E}_1$ describes the tissue mechanics through cell area and perimeter elasticity:
\begin{equation}
\mathcal{E}_1 = \sum_{\text{cell }\alpha} \left[ K_{A,\alpha} \left( A_\alpha - A^0_\alpha \right)^2 + K_{P,\alpha} \left( P_\alpha - P^0_\alpha \right)^2 \right],
\end{equation}
where for each cell $\alpha$, $A_\alpha$ and $P_\alpha$ are its actual area and perimeter, $A^0_\alpha$ and $P^0_\alpha$ are the target area and perimeter, and $K_{A,\alpha}$ and $K_{P,\alpha}$ are the associated elastic stiffness moduli. To simplify the parameter space, this energy is often non-dimensionalized, leading to the definition of a target shape factor $p^0_\alpha = P^0_\alpha / \sqrt{A^0_\alpha}$.

Alternatively, \texttt{VertAX} implements the formulation $\mathcal{E}_2$, which explicitly accounts for distinct physical contributions from cell-cell adhesion and cortical tension:
\begin{equation}
\mathcal{E}_2 = \sum_{\text{cell} \alpha} \frac{1}{2} K \left( A_\alpha - A^0_\alpha \right)^2 + \sum_{\text{edge }\langle i,j \rangle} \gamma_{ij} \ell_{ij},
\end{equation}
where $\gamma_{ij}$ denotes the line tension along the edge connecting vertices $i$ and $j$, and $\ell_{ij}$ is the length of that edge.

\subsection*{System state and data structure}

The state of the system is fully defined by the two-dimensional spatial coordinates of the $N$ vertices, represented by the vector $\mathbf{X} \in \mathbb{R}^{2N}$. In the following the \textit{State Size} is $2N$. The physical properties of the tissue are encapsulated in a parameter vector $\boldsymbol{\theta} \in \mathbb{R}^M$, which can include shape factors, elastic moduli, or line tensions.

To allow for efficient, vectorized operations during JAX compilation, the tissue graph is encoded using a Half-Edge data structure split into distinct tables:
\begin{itemize}
    \item \texttt{vertTable}: Stores the spatial geometry (the $x$ and $y$ coordinates of each vertex).
    \item \texttt{heTable}: Stores the topological connectivity. It contains eight columns: the indices of the previous, next, and twin half-edges; the indices of the source and target vertices; the index of the face to which the half-edge belongs; and finally, two columns storing offset information ($+1$, $0$, or $-1$) for the target vertex along the $x$ and $y$ axes to handle periodic boundary conditions.
    \item \texttt{faceTable}: Stores cell-level properties and topology. For specific inverse design tasks (e.g., cell fate patterning), an extra column is added to encode target fate information.
\end{itemize}

\subsection*{Boundary conditions, Topology changes and Initialization}

\paragraph{Periodic mode.} To simulate bulk tissue dynamics without boundary effects, the tissue is enclosed within a two-dimensional simulation box with periodic boundary conditions. In addition to the primary region, eight neighboring (mirrored) regions are considered. Half-edges that cross the boundary of the primary region are split: the source vertex remains within the primary box, while the target vertex is assigned a spatial offset based on its destination region (stored in the \texttt{heTable}). Vertex positions and their corresponding target offsets are updated at each optimization step.

\paragraph{Bounded mode.} To simulate finite tissue clusters (e.g., organoids or isolated monolayers) interacting with an external medium, \texttt{VertAX} implements a bounded mode. Edges in contact with the external medium (those belonging to only a single cell face) are modeled as circular arcs rather than straight lines, allowing for geometric rounding at the boundary. This introduces an additional physical degree of freedom for each boundary edge: the angle $\alpha$ subtending the arc. These angles are stored in an additional \texttt{angTable} and are optimized simultaneously with the vertex positions to satisfy mechanical equilibrium at the outer junctions.

\paragraph{T1 transitions.} Epithelial tissues undergo topological rearrangements known as T1 transitions, where cells exchange neighbors. During each iteration, edges are sequentially checked. If an edge length falls below a user-defined threshold $\ell_{\text{min}}$, two candidate topological updates are evaluated:
1. The edge is stretched along its current orientation.
2. A T1 transition is executed by deleting the short edge and creating a new orthogonal edge.
The configuration that yields the lower total energy is accepted.

\paragraph{Mesh initialization.} Initial configurations can be generated using two dedicated routines. The first constructs a periodic Voronoi tessellation from user-defined seed points using \texttt{scipy.spatial.Voronoi}. To ensure periodicity, the seed configuration is replicated across the eight neighboring regions before computing the tessellation, after which only the polygons within the primary region are retained. Boundary half-edges are subsequently assigned the correct topological offsets. Alternatively, meshes can be initialized directly from microscopy images using segmentation tools (e.g., Cellpose). To enforce periodicity in image-derived meshes, the image is mirrored into the neighboring regions prior to segmentation. In the bounded mode, initialization follows the same Voronoi procedure, but boundary edges are simply identified and assigned a random initial angle between $0$ and $\pi/2$.

\subsection*{Simulation engine}

The physical relaxation of the tissue is simulated by finding the vertex configuration that minimizes the energy functional for a given set of parameters. This inner optimization is performed using gradient descent (GD):
\begin{equation}
\mathbf{X}_{t+1} = \mathbf{X}_t - \eta_{\text{inner}} \nabla_{\mathbf{X}} \mathcal{E}(\mathbf{X}_t, \boldsymbol{\theta}),
\end{equation}
where $\eta_{\text{inner}}$ is the step size.

The inner energy minimization loop halts when the maximum nodal force falls below a specified tolerance ($\max(|\nabla_{\mathbf{X}} \mathcal{E}|) < \epsilon$) over a defined patience window, or when a maximum number of iterations is reached. The resulting equilibrium configuration is denoted as $\mathbf{X}^*_{\boldsymbol{\theta}}$\footnote{The configuration $\mathbf{X}^{*}_\theta$ represents a true stable mechanical equilibrium when the corresponding Hessian, $\mathbf{H}_{\mathbf{X}} = \nabla^2_{\mathbf{X}}\mathcal{E}(\mathbf{X}^{*}_\theta,\boldsymbol{\theta})$, is positive definite (\emph{i.e.} $\boldsymbol{v}^\mathsf{T}\mathbf{H}_{\mathbf{X}}\boldsymbol{v} > 0$ for all nonzero $\boldsymbol{v}$).}.

\subsection*{Bilevel optimization framework}

To infer mechanical parameters or design tissue configurations, \texttt{VertAX} employs a bilevel optimization (BO) architecture. The objective is to find the optimal parameter set $\boldsymbol{\theta}^*$ that minimizes an outer cost function $\mathcal{C}$, evaluated at the equilibrium state $\mathbf{X}^*_{\boldsymbol{\theta}}$:
\begin{equation}
\boldsymbol{\theta}^* = \arg\min_{\boldsymbol{\theta}} \mathcal{C}(\mathbf{X}^*_{\boldsymbol{\theta}}, \theta),
\end{equation}
subject to the inner constraint:
\begin{equation}
\mathbf{X}^*_{\boldsymbol{\theta}} \in \arg\min_{\mathbf{X}} \mathcal{E}(\mathbf{X}, \boldsymbol{\theta}).
\end{equation}
The outer parameter update at each epoch $k$ is computed using the ADAM optimizer:
\begin{equation}
\theta_{k+1} = \mathrm{ADAM}(\theta_k, \nabla_\theta C, \eta_{\mathrm{outer}}) 
\end{equation}
where $\eta_{\mathrm{outer}}$ is the learning rate.

\subsection*{Gradient computation methods}

Solving the BO problem requires computing the hypergradient $\nabla_{\boldsymbol{\theta}} \mathcal{C}$, which involves differentiating through the inner energy minimization process. \texttt{VertAX} supports three distinct strategies for computing this hypergradient.

\subsubsection*{Automatic differentiation (AD)}
AD computes the exact hypergradient $\nabla_{\boldsymbol{\theta}} \mathcal{C}$ by differentiating through the full sequence of operations performed during the inner minimization loop, using reverse-mode automatic differentiation~\cite{baydin2018automatic}. This requires storing the entire unrolled computational graph in memory, resulting in a memory complexity of $\mathcal{O}(N_{\text{iter}} \times \text{State Size})$.

In practice, implementing reverse-mode AD in JAX imposes strict structural constraints on the vertex model. The inner energy minimization cannot rely on dynamic control flow such as \texttt{lax.while\_loop}, which discards intermediate states for memory efficiency and is therefore incompatible with reverse-mode AD primitives such as \texttt{jax.grad} or \texttt{jax.jacrev}. Instead, the relaxation dynamics must be unrolled over a fixed number of iterations using \texttt{lax.scan}, which preserves the full primal trajectory in a static computational graph amenable to backpropagation. Likewise, neighborhood queries and local topological operations must be padded to a fixed maximum size so that all tensor dimensions are known at trace time. Under these conditions, reverse-mode AD can backpropagate the scalar outer loss through the iterative physics solver and yields exact parameter gradients, making it particularly effective when the number of parameters greatly exceeds the number of objective function outputs.

\subsubsection*{Implicit differentiation (ID)}
ID exploits the Implicit Function Theorem at the equilibrium state, where the net physical force on all vertices is zero: $\nabla_{\mathbf{X}} E(\mathbf{X}^*_{\boldsymbol{\theta}}, \boldsymbol{\theta}) = 0$. Differentiating this optimality condition yields the exact hypergradient:
\begin{equation}
\nabla_{\boldsymbol{\theta}} \mathcal{C} = \partial_{\boldsymbol{\theta}} \mathcal{C} - \left(\nabla_{\mathbf X \boldsymbol{\theta}}^2 E\right)^{\!\top} \:\mathbf H_{\mathbf X}^{-1} \:\nabla_{\mathbf X} \mathcal{C},
\end{equation}
where $\mathbf H_{\mathbf X} = \nabla_{\mathbf X}^2 E$ is the Hessian matrix of the inner energy. By evaluating this equation solely at the final equilibrium, ID avoids loop unrolling and reduces the memory footprint to $\mathcal{O}(\text{State Size})$.

ID can be implemented in two mathematically equivalent but computationally distinct ways. The forward mode, or Sensitivity ID (Sen-ID), computes Jacobian–vector products to directly propagate the sensitivities of the equilibrium state $\mathbf{X}^*_{\boldsymbol{\theta}}$ with respect to the parameters $\boldsymbol{\theta}$. This approach requires solving one linear system per parameter direction. Alternatively, the reverse mode, or Adjoint-state ID (Adj-ID)~\cite{plessix2006review}, computes vector–Jacobian products to evaluate the sensitivity of the outer loss with respect to the inner variables, requiring only a single backward linear solve regardless of the parameter space dimension.

Theoretical formulations of ID assume an invertible inner Hessian~\cite{dontchev2009implicit, zucchet2022beyond}. However, mechanical equilibria in vertex models often do not correspond to strictly stable minima. Marginal modes arising from boundary degrees of freedom, recent topological rewiring (T1 transitions), or incomplete energy relaxation can result in $\mathbf H_{\mathbf X}$ being indefinite. Consequently, direct linear solvers may yield unstable or divergent gradients. To robustly handle these indefinite operators, \texttt{VertAX} solves the ID linear systems using Krylov subspace methods, specifically \texttt{minres}.

\subsubsection*{Equilibrium propagation (EP)}
EP estimates the bilevel gradient by introducing a small perturbation to the energy landscape~\cite{scellier2017equilibrium}. The physical energy is augmented with the outer cost function:
\begin{equation}
E_{\text{aug}}(\mathbf{X}, \boldsymbol{\theta}, \beta) = E(\mathbf{X}, \boldsymbol{\theta}) + \beta \mathcal{C}(\mathbf{X}, \mathbf{X}_{\text{target}}),
\end{equation}
where $0 < \beta \ll 1$ is a small scalar that biases the system toward reducing the outer loss. The system is allowed to relax to a new ``nudged'' equilibrium state, $\mathbf{X}^*_{\boldsymbol{\theta}, \beta}$, that minimizes $E_{\text{aug}}$.

In a standard first-order formulation, the EP gradient estimator is computed by comparing the response of the nudged state to the free (unperturbed) equilibrium $\mathbf{X}^*_{\boldsymbol{\theta}, 0}$:
\begin{equation}
\nabla_{\boldsymbol{\theta}} \mathcal{C} \approx \frac{1}{\beta} \left[ \nabla_{\boldsymbol{\theta}} E_{\text{aug}}(\mathbf{X}^*_{\boldsymbol{\theta}, \beta}, \boldsymbol{\theta}) - \nabla_{\boldsymbol{\theta}} E_{\text{aug}}(\mathbf{X}^*_{\boldsymbol{\theta}, 0}, \boldsymbol{\theta}) \right].
\end{equation}

To achieve higher accuracy and cancel leading-order approximation errors~\cite{laborieux2021scaling}, \texttt{VertAX} natively implements a centered (second-order) finite-difference estimator using symmetric perturbations at $\pm \beta/2$:
\begin{equation}
\nabla_{\boldsymbol{\theta}} \mathcal{C} \approx \frac{1}{\beta} \left[ \nabla_{\boldsymbol{\theta}} E(\mathbf{X}^*_{\boldsymbol{\theta}, +\beta/2}, \boldsymbol{\theta}) - \nabla_{\boldsymbol{\theta}} E(\mathbf{X}^*_{\boldsymbol{\theta}, -\beta/2}, \boldsymbol{\theta}) \right].
\end{equation}
This centered estimator yields more accurate and stable gradients in practice, making it the preferred EP method in our framework~\cite{scellier2023energy}.

\subsection*{Outer loss functions}

\subsubsection*{Geometric vertex-to-vertex (v2v) cost}
The v2v cost measures the direct spatial deviation between the predicted vertices $\mathbf{X}$ and the target vertices $\mathbf{X}_{\text{target}}$ using the mean squared error (MSE):
\begin{equation}
\mathcal{C}_{\text{v2v}} = \frac{1}{2N} \sum_{i=1}^{N} \|\mathbf{x}_i - \mathbf{x}_{\text{target},i}\|^2.
\end{equation}

\subsubsection*{Index-Aware Structural (IAS) cost}
To provide a differentiable pathway for topological rewiring, we define a continuous structure matrix $S \in \mathbb{R}^{N \times N}$ based on the dual adjacency matrix $A$ of the tissue:
\begin{equation}
    S = 2I - A - \frac{1}{2}A^2.
\end{equation}
The structural cost matrix $C_{\text{struct}} \in \mathbb{R}^{N \times N}$ between the predicted and target tissues is defined by the row-wise $L^2$ norm of their respective structure matrices:
\begin{equation}
    [C_{\text{struct}}]_{ij} = \left\| S^{(1)}_{i, \cdot} - S^{(2)}_{j, \cdot} \right\|_2.
\end{equation}
Driven by this structural discrepancy, the topological matching is reduced to finding the optimal transport plan $P$ that minimizes the entropy-regularized Kantorovich problem:
\begin{equation}
    \mathcal{L}_{\text{IAS}} = \min_{P \in \Pi(\mathbf{a}, \mathbf{b})} \langle P, C_{\text{struct}} \rangle_F + \varepsilon H(P),
\end{equation}
where $\Pi(\mathbf{a}, \mathbf{b})$ is the transportation polytope for uniform marginals $\mathbf{a}=\mathbf{b}=\frac{1}{N}\mathbf{1}$, $\langle \cdot, \cdot \rangle_F$ is the Frobenius inner product, and $H(P)$ is the entropic regularization term. This is solved efficiently using the Sinkhorn-Knopp algorithm~\cite{cuturi2013sinkhorn}.

\subsubsection*{$L^2$ Mesh-Image cost}
To compare a simulated mesh $\mathbf{X}$ directly to a microscopy image $I$, the mesh is formalized as a compact measure $\mu$ on $\mathbb{R}^2$ (the sum of Dirac measures on each line segment). The mesh is transformed into a continuous simulated image by convolving $\mu$ with a Gaussian point spread function (PSF) $g$, with variance $\sigma^2$. The image $I$ and the blurred mesh $\mu*g$ are compared using the $L^2$ norm, which is computed efficiently in the Fourier domain: $\|I-\mu*g\|_{L^2} = \|\hat I-\hat\mu\cdot\hat g\|_{L^2}/(2\pi)$. 

For a generic line segment $S$ with length $|S|$, midpoint $\boldsymbol{m}$, and half-directional vector $\boldsymbol{d}$, the Fourier transform of its Dirac measure $\mu_S$ evaluates analytically to:
\[
\hat\mu_S(\boldsymbol{\xi})=|S|\mathrm{e}^{-\mathrm{i}\boldsymbol{\xi}\cdot\boldsymbol{m}}j_0(\boldsymbol{\xi}\cdot\boldsymbol{d})\quad,\quad\boldsymbol{\xi}\in\mathbb{R}^2,
\]
where $j_0(t)=\sin(t)/t$ is the unnormalized spherical Bessel function. By summing $\hat\mu_S$ over all line segments, the analytical Fourier transform of the entire mesh $\hat\mu_{\boldsymbol{\bar X}}$ is computed and multiplied by the Gaussian $\hat g(\boldsymbol{\xi})=\mathrm{e}^{-\sigma^2|\boldsymbol{\xi}|^2/2}$ to execute the differentiable image-mesh cost, similar to its 3D counterpart on triangular meshes~\cite{ichbiah2025inverse}.

\subsection*{Experimental setup}

\subsubsection*{Cell culture and \textit{in vitro} model}
Human lung carcinoma epithelial cells (A549, ATCC) were cultured in Dulbecco’s Modified Eagle Medium (DMEM) supplemented with 10\% fetal bovine serum (FBS) and 1\% penicillin-streptomycin. Cells were maintained in a humidified incubator at 37°C with 5\% CO$_2$. For imaging experiments, cells were seeded in 35 mm glass-bottom dishes (Cellvis) and allowed to adhere overnight. To induce Epithelial-Mesenchymal Transition (EMT), the culture medium was replaced with DMEM containing 5 ng/ml of Recombinant Human TGF-$\beta$1 (PeproTech) for 48 hours prior to imaging. Control samples (CTRL) were maintained in standard medium. 

\subsubsection*{Immunostaining}
Cells were grown to full confluency prior to staining. The culture medium was aspirated, and the monolayers were gently washed with 1 ml of Phosphate-Buffered Saline (PBS). Live-cell staining for E-cadherin was performed by incubating the cells with an APC-conjugated E-cadherin antibody (Thermo Fisher Scientific, MA1-10193) diluted in 750 $\mu$l of PBS (5 $\mu$l antibody per dish) for 30 minutes at 4°C. Following incubation, the cells were washed with 1 ml of PBS and subsequently counterstained with DAPI (Thermo Fisher Scientific, 62248; 1:1000 dilution in 750 $\mu$l of PBS) for 5 minutes at room temperature. After a final PBS wash, the dishes were mounted using Ibidi Mounting Medium (50001) to preserve fluorescence prior to imaging.

\subsubsection*{Spinning Disk Microscopy}
Brightfield and widefield fluorescence images were acquired using an inverted Olympus iX73 microscope equipped with an X-Light V3 spinning disk confocal head (Crest Optics), an LDI laser illuminator (89 North), and a Prime BSI sCMOS camera (Photometrics). Fluorescence excitation was achieved using 405 nm and 640 nm lasers for DAPI and APC, respectively. Image acquisition was performed using a 20× objective (Olympus) and controlled via MetaMorph software (Molecular Devices). Images were subsequently processed and analyzed using ImageJ and custom Python scripts.

\end{document}